\documentclass[runningheads]{llncs}
\usepackage[T1]{fontenc}
\usepackage{graphicx}
\usepackage{amsmath}
\usepackage{amsfonts}

\usepackage{algorithm}
\usepackage{algpseudocode}
\algnewcommand{\Inputs}[1]{%
  \State \textbf{Inputs:}\hspace*{\algorithmicindent}\parbox[t]{.8\linewidth}{\raggedright #1}
}
\algnewcommand{\Initialize}[1]{%
  \State \textbf{Initialize:} \hspace*{\algorithmicindent}\parbox[t]{.8\linewidth}{\raggedright #1}
}

\begin{document}
%
%
\title{Collective Bayesian Decision-Making in a Swarm of Miniaturized Robots for Surface Inspection}
%
%
\titlerunning{Collective Bayesian Decision-Making for Surface Inspection}
%
%
\author{Thiemen Siemensma\inst{1} \and
Darren Chiu\inst{2} \and
Sneha Ramshanker\inst{3} \and 
Radhika Nagpal \inst{3} \and
Bahar Haghighat \inst{1}}
%
\authorrunning{T. Siemensma et al.}
%
\institute{University of Groningen, Groningen, The Netherlands \email{\{t.j.j.siemensma,bahar.haghighat\}@rug.nl} \and University of Southern California, Los Angeles, CA, USA
\email{chiudarr@usc.edu}\\ \and
 Princeton University, Princeton, NJ, USA \email{\{sr6848,rn1627\}@princeton.edu}}
%
\index{Siemeensma Thiemen}
\index{SurnameAuthor2, FirstnameAuthor2}
\index{SurnameAuthor3, FirstnameAuthor3}
%
\maketitle              
\begin{abstract}
Robot swarms can effectively serve a variety of sensing and inspection applications. Certain inspection tasks require a binary classification decision. This work presents an experimental setup for a surface inspection task based on vibration sensing and studies a Bayesian two-outcome decision-making algorithm in a swarm of miniaturized wheeled robots. The robots are tasked with individually inspecting and collectively classifying a $1m \times 1m$ tiled surface consisting of vibrating and non-vibrating tiles based on the majority type of tiles. The robots sense vibrations using onboard IMUs and perform collision avoidance using a set of IR sensors. We develop a simulation and optimization framework leveraging the Webots robotic simulator and a Particle Swarm Optimization (PSO) method. We consider two existing information sharing strategies and propose a new one that
allows the swarm to rapidly reach accurate classification decisions. We first find optimal parameters that allow efficient sampling in simulation and then evaluate our proposed strategy against the two existing ones using 100 randomized simulation and 10 real experiments. We find that our proposed method compels the swarm to make decisions at an accelerated rate, with an improvement of up to 20.52\% in mean decision time at only 0.78\% loss in accuracy.
\end{abstract}
\section{Introduction}
Over the last few decades, automated inspection systems have increasingly become a valuable tool across various industries \cite{Brem2023Senseye2022,PwC2017PdM4.0,Roda2018TheManufacturing,Bousdekis2020PredictiveImplications}. Studies have addressed applications in agricultural and hull inspection as well as infrastructure and wind-turbine maintenance \cite{Carbone2018SwarmAgriculture,Liu2022ReviewTurbines,Lee2023SurveyInspection,Halder2023RobotsReview}. Vibration analysis is a valuable tool in these inspection processes. Different types of vibration analysis are used to detect the condition of infrastructure through structural properties such as modal shapes and eigenfrequencies \cite{ArnaudDeraemaeker2010NewMonitoring,Magalhaes2012VibrationDetection,Doebling1996DamageReview}. A class of inspection tasks involves making a binary decision about a spatially distributed feature of the inspected system. This type of decisions can be effectively addressed by a swarm of robots \cite{Valentini2017ThePerspectives,Schranz2020SwarmApplications}. When compared to a single entity, swarms improve decision time and accuracy by leveraging collective perception \cite{Valentini2016CollectiveSwarm,Valentini2016CollectiveSystems}. Moreover, swarms eliminate the problem of sensor-placement and can provide a high-resolution map of the environment \cite{Bayat2017EnvironmentalTechniques,Bigoni2020SystematicStates}. Collective decision-making algorithms often draw inspiration from nature, such as groups of ants and bees \cite{Seeley1999GroupBees}. A more mathematical approach is found in Bayesian algorithms. Applications of Bayesian algorithms have been studied in sensor networks \cite{Alanyali2004DistributedNetworks,Makarenko2006DecentralizedNetworks} as well as robot swarms \cite{Valentini2016CollectiveSwarm,Ebert2018Multi-FeatureSwarms,Valentini2016CollectiveSystems,Ebert2020BayesSwarms,Haghighat2022AOrbit}. 
In the study outlined in \cite{Ebert2020BayesSwarms}, a collective of agents must determine whether the predominant color of a checkered surface pattern is black or white. The robots function as Bayesian modellers, exchanging information based on two information sharing strategies. The robots either (i) continuously broadcast their ongoing binary observations (\textit{no feedback}) or (ii) continuously broadcast their irreversible decisions once reached (\textit{positive feedback}), pushing the swarm to consensus. A common problem with collective perception methods is slow convergence, i.e. difficulty reaching high belief (probability) about the predominant color. Spatial correlation of observations cause the belief to fluctuate, resulting in long decision times and low decision accuracies, as shown by \cite{Ebert2018Multi-FeatureSwarms,Bartashevich2021Multi-featuredCorrelations,Bartashevich2019BenchmarkingDecision-making}. 

In this work, we build on top of the work in \cite{Ebert2020BayesSwarms} in two ways. First, we propose a novel information sharing strategy, named \textit{soft-feedback}. In this approach, binary information is shared between the robots based on their current belief, similar to \cite{Shan2021DiscreteSharing,Bartashevich2021Multi-featuredCorrelations}, with the addition of sample variance and random sampling. The resulting strategy is shown to enhance convergence compared to no feedback and positive feedback, without compromising the accuracy of decisions. Second, we move away from the agent-based simulation setup of \cite{Ebert2020BayesSwarms} and present a real experimental setup built around 3-cm-sized vibration sensing wheeled robots. We utilize vibration signals in the presence of measurement noise in place of simulated binary floor color observations. We develop a new sensor board to allow the robots to perform collision avoidance. A digital twin of our experimental setup is developed in Webots, derived from the work done in \cite{Chiu2023OptimizationOptimization}. We use this model for calibration of our robots and optimization of the algorithm parameters in a Particle Swarm Optimization (PSO) loop.

\section{Problem Definition}
We task a swarm of $N$ robots to individually inspect and collectively classify a 2D tiled surface section. The surface comprises two types of tiles, vibrating and non-vibrating tiles. The swarm must determine whether the tiled surface is majority vibrating or majority non-vibrating. We denote the fill-ratio $f$ as the proportion of vibrating tiles. Thus, fill-ratios close to $0.5$ represent a hard surface inspection problem, as the amount of vibrating and non-vibrating tiles is almost equal.
The robots each individually inspect the surface, share their information with the rest of the swarm, and collectively classify the surface. The inspection task ends when every robot in the swarm reaches a final decision, determining if the fill-ratio is above or below 0.5. An underlying real-world scenario could involve inspecting a surface section and determining whether the surface is in a majority \textit{healthy} or in a majority \textit{unhealthy} state.
\begin{figure}[t]
    \centering
    \begin{minipage}[b]{0.45\textwidth}
        \centering
        \includegraphics[width=\textwidth]{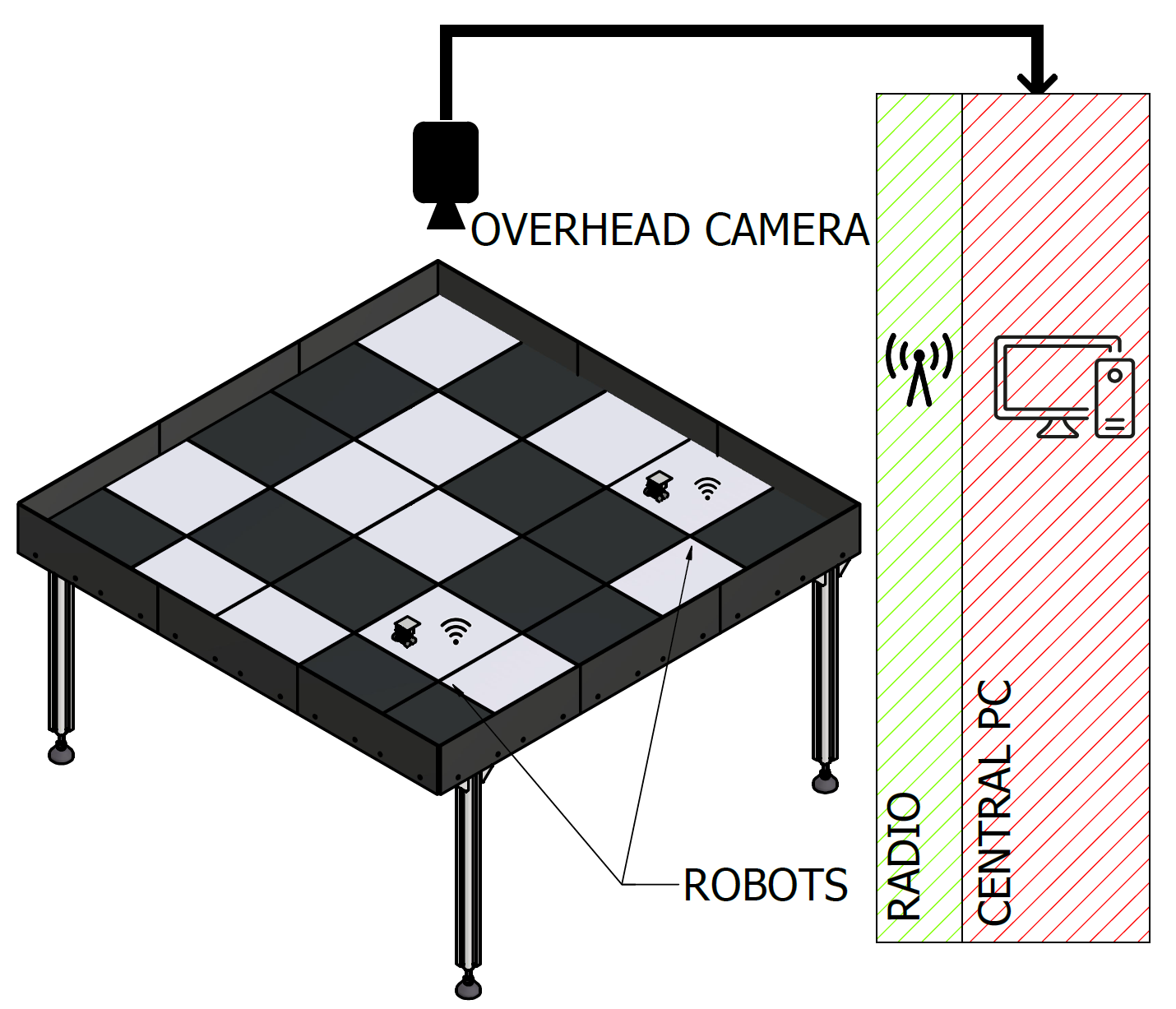}
        (a) Overall experimental setup. 
    \end{minipage}
    \hspace{0.05\textwidth}
    \begin{minipage}[b]{0.4\textwidth}
        \centering
        \includegraphics[width=\textwidth]{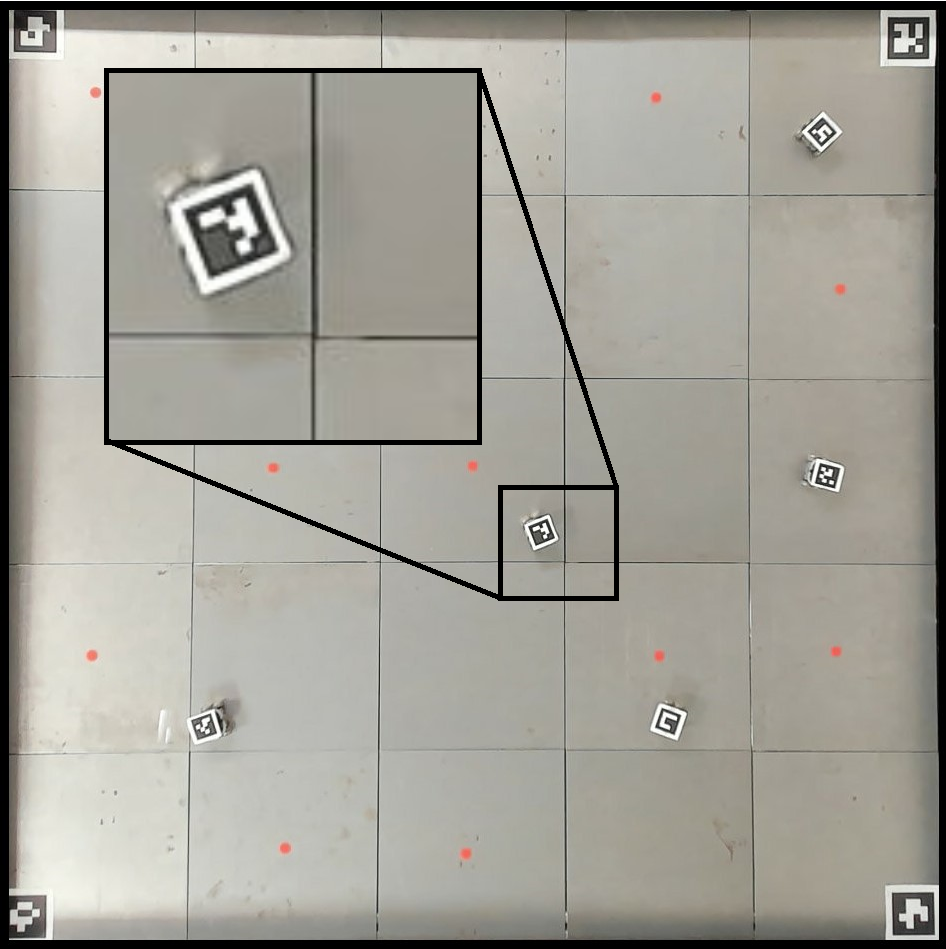}
        (b) View from overhead camera.
    \end{minipage}
\caption{The experimental setup with a fill ratio of $f = \frac{12}{25} = 0.48$. (a) Schematic overview of the setup is shown. The central PC uses the radio and camera data for analysis. Vibration-motors are attached on the bottom side of white tiles. (b) A snapshot from the overhead camera with detailed view (black square). The red dot markings indicate vibrating tiles. Each robot carries a unique AruCo marker for tracking. AruCo markers in the corners of the environment mark the boundaries.}
\label{fig:experimental_setup}
\end{figure}
\section{Inspection Algorithm}
\label{sec:algorithm}
Algorithm \ref{alg:updatealgo} shows our collective Bayesian decision making algorithm. The robots individually estimate and classify the fill ratio $f$ as above or below 0.5.
Each robot acts as a Bayesian modeler integrating personal observations and information broadcast by other robots.  We consider three information sharing strategies: the (i) no feedback ($u^-$) and (ii) positive feedback ($u^+$) strategies, previously studied in \cite{Ebert2020BayesSwarms}, and the (iii) soft feedback ($u^s$) strategy, which we propose in this work. 

The robots make binary observations of the surface condition as black/white in simulation or vibrating/non-vibrating in the real setup as $O \in \{0,1\}$:
\begin{equation}
    O \sim \textrm{Bernoulli}(f)
\end{equation}
The fill ratio $f \in [0,1]$ is unknown and is modeled by a Beta-distribution: 
\begin{equation}
    f \sim \textrm{Beta}(\alpha,\beta)
\end{equation}
The prior distribution of $f$ is initialized as $\text{Beta}(\alpha_0 = 1,\beta_0=1)$. Upon sampling or receiving observations from other robots, the posterior of $f$ is updated as:
\begin{equation}
    f \ | \ O \sim \textrm{Beta}(\alpha + O, \beta+ (1- O))
\end{equation}
\begin{algorithm}[t]
\caption{Collective Bayesian Decision Making}\label{alg:updatealgo}
\begin{algorithmic}
\Inputs{$u^-,u^+,u^s, T_{end},\theta_o,\eta,\theta_c,\tau,p_c$}
\Initialize{ $\alpha = 1, \beta =1,d_f =-1, \text{robot } id ,t_s =0$}
\While{$t < T_{end}$}
\State Perform random walk for $\tau$ time
\If{$t - t_s > \tau$}
    \State $O\gets $ Observation\Comment{Get binary observation}
    \State $t_s\gets t$ \Comment{Observation timestamp}
    \State $\text{Beta}(\alpha,\beta) \gets \textrm{Beta}(\alpha + O, \beta+ (1- O)) $ \Comment{Update modeling of $f$}
    \State $p \gets P(\text{Beta}(\alpha,\beta)<0.5) $\Comment{Update belief on $f$}
    \State $O_{count} \gets O_{count}+1 $ \Comment{Observation count}
\EndIf
\If{$(u^s \text{ \textbf{Or} } d_f == -1)$  \textbf{And} $ (O_{count} > \theta_o)$}
    \If{$p>p_c$}
        \State $d_f \gets 0$
    \ElsIf{$(1-p) > p_c $}
     \State $d_f \gets 1$
    \EndIf
\EndIf
\If{$u^s$}
 \State $\Gamma \gets \text{Var(Beta)}$
 \State $ m\leftarrow \text{Bernoulli}\left((1-p)e^{-\eta \Gamma} (\frac{1}{2} - p)^2 + O(1 - e^{-\eta \Gamma}(\frac{1}{2} - p)^2)\right)$
 \State Broadcast($m$) \Comment{Soft feedback}
\ElsIf{($u^+$ \textbf{And} $d_f \neq -1$)}
 \State Broadcast($d_f$) \Comment{Positive feedback}
 \Else 
 \State Broadcast($O$)\Comment{No feedback}
\EndIf
\If{Message in queue}
    \State $O \gets $ Message \Comment{Receive message from swarm}
    \State $\text{Beta}(\alpha,\beta) \gets \textrm{Beta}(\alpha + O, \beta+ (1- O))$ \Comment{Update modeling of $f$}
\EndIf
\EndWhile
\end{algorithmic}
\end{algorithm}
The robots perform a Levy-flight type random walk, moving forward for a time drawn from a Cauchy distribution with mean $\gamma_0$ and average absolute deviation $\gamma$ followed by turning a uniform random angle $\phi \sim U(-\pi,\pi)$ in the direction of $\text{sign}(\phi)$ relative to the forward driving direction. The robots perform collision avoidance upon detecting an obstacle within a range of $\theta_c$ millimeters, by turning a random angle $\phi \sim U(-\pi,\pi)$ in the direction of $\text{sign}(\phi)$ relative to the forward driving direction.

Every $\tau$ milliseconds a robot samples a new observation $O$. In simulation, $O$ is based on a binary floor color sampling. In experiments, $O$ is calculated using a 500 millisecond vibration signal sample. The DC component of this sample is removed by employing a first-order high-pass filter with cutoff frequency $\omega_n = 40$ Hz. Given a sampling rate of $350$ Hz, the filter parameters $\alpha_1$, $\alpha_2$, and $\alpha_3$ are configured to values: $0.20$, $0.60$, and $-0.60$ respectively. We define the filtered signal at time step $i$ as $\hat a_i$:
\begin{equation}
    \hat a_i :=  \alpha_1 \hat a_{i-1} +\alpha_2 a_i  +\alpha_3 a_{i-1}
\end{equation}
where $a_i$ is the magnitude of the IMU's raw acceleration data $a_x$, $a_y$, and $a_z$. The Root-Mean-Square (RMS) of $\hat a$ returns the energy of the signal as $\hat E =\sqrt{\frac{1}{n}\sum_{i=1}^n \hat a_i^2}$. Subsequently, the observation $O$ is determined by comparing $\hat E$ with a threshold $\theta_E$:
\begin{equation}
\label{eq:observation}
    O=\begin{cases}
            1& \text{if } \hat E >\theta_E\\
             0& \text{if } \hat E \leq \theta_E
        \end{cases}\\
\end{equation}
We consider three information sharing strategies. (i) No feedback ($u^-$) considers sharing the latest observation $O$ in any case. (ii) Positive feedback ($u^+$) is similar to ($u^-$) until reaching a final decision. From this point in time it broadcasts its irreversible final decision $d_f$. The intuition behind positive feedback is to push the swarm to consensus upon reaching a final decision by one robot. However, this is not very effective when no robot is able to reach a final decision. In this case, positive feedback is not different from no feedback. To resolve this, we propose soft-feedback (iii). Soft feedback ($u^s$) broadcasts a binary value sampled from a Bernoulli distribution. The corresponding probability is calculated through the soft feedback parameter $\eta \in \mathbb{R}^+$, the current observation $O \in \{0,1\}$, and the current belief $p \in [0,1]$ as below:
\begin{subequations}
\begin{equation}    
    \label{eq:soft_feedback}
            m \sim \text{Bernoulli}\left(\delta \cdot (1-p) + (1 - \delta) \cdot O\right)
\end{equation}
\begin{equation}
    \label{eq:delta}
    \delta = e^{-\eta \Gamma} (\frac{1}{2} - p)^2
\end{equation} 
\begin{equation}
    \label{eq:belief}
    p = P(\text{Beta}(\alpha,\beta) < 0.5)  
\end{equation} 
\end{subequations}
where $m\in \{0,1\}$ is the outgoing message, $\Gamma$ is the variance of the Beta distribution and $p$ is the robot's belief evaluated as the CDF of the Beta distribution at $f=0.5$. The intuition behind soft-feedback is to broadcast information that is initially incorporating only the observation $O$, but gradually factors in more of the belief $p$. Namely, Equation \ref{eq:delta} depends on (i) a compelling component $e^{-\eta \Gamma} \in [0,1]$ which increases the proportion of the current belief in messages as $\Gamma$ decreases, and (ii) a stabilizing component $(\frac{1}{2}-p)^2 \in [0,0.25]$ which is the squared distance of $p$ from the indecisive state $p=0.5$, increasing the proportion of robot's belief in the Bernoulli sampled message to enhance accuracy. This prevents the robot from prematurely making a decision with low confidence.

Upon reaching a minimum of $\theta_o$ number of observations, a robot considers making a final decision based on its belief $p$. If above the credibility threshold $p>p_c$, the robot's final decision $d_f$ is set to 0. Conversely, if $(1-p)>p_c$, $d_f$ is set to 1. The inspection task ends when all robots have made a final decision.

\begin{figure}[t!]
    \centering
    \begin{minipage}[t]{0.2\textwidth}
        \centering
        \includegraphics[width=\textwidth]{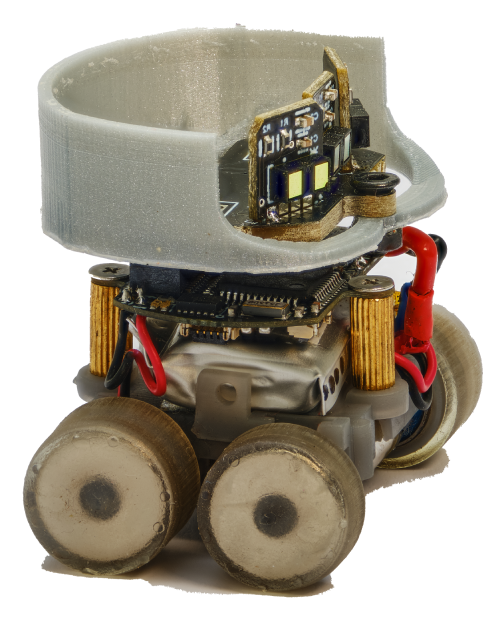}
        (a) Real Robot. 
    \end{minipage}
    \begin{minipage}[t]{0.4\textwidth}
        \centering
        \includegraphics[width=\textwidth]{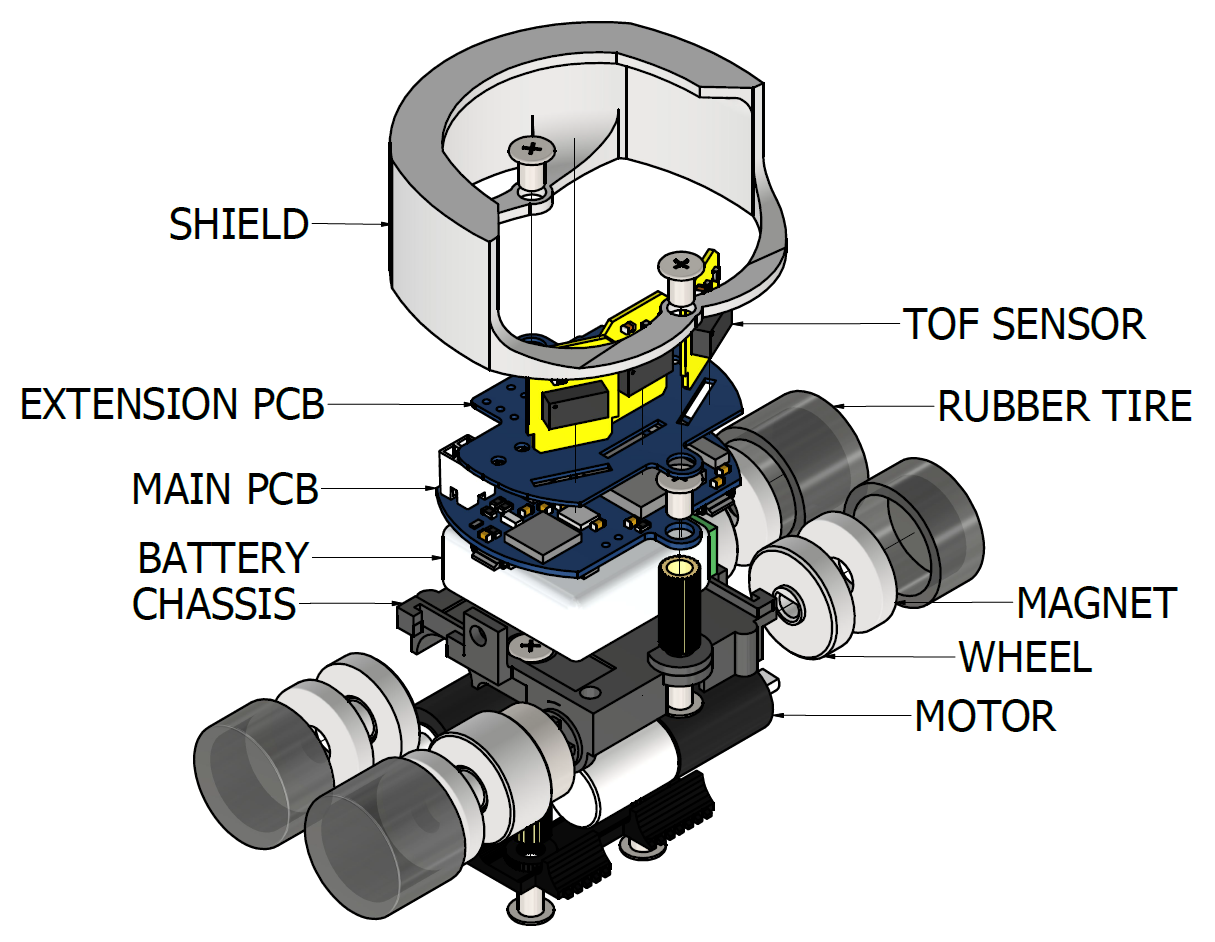}
        (b) Robot CAD view
    \end{minipage}
    \begin{minipage}[t]{0.225\textwidth}
        \centering
        \includegraphics[width=\textwidth]{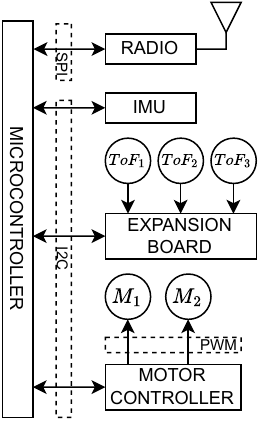}
        (c) Schematics
    \end{minipage}
\caption{We use a revised and extended version of the original Rovable robot [9].
(a) The extended robot with IR sensor board. (b) Exploded 3D CAD view of the
extended robot. (c) Electronic block diagram of the extended robot. The microcontroller (Atmel SAMD21G18) interfaces with the IMU (MPU6050), 2.4 GHz radio (nRF24L01+), motor-controllers (DRV8835), and ToF IR sensors (VL53L1X).}
\label{fig:vibration_sensing_robot}
\end{figure}
\begin{figure}[t!]
    \centering
    \begin{minipage}[t]{0.19\textwidth}
        \centering
        \includegraphics[width=0.9\textwidth]{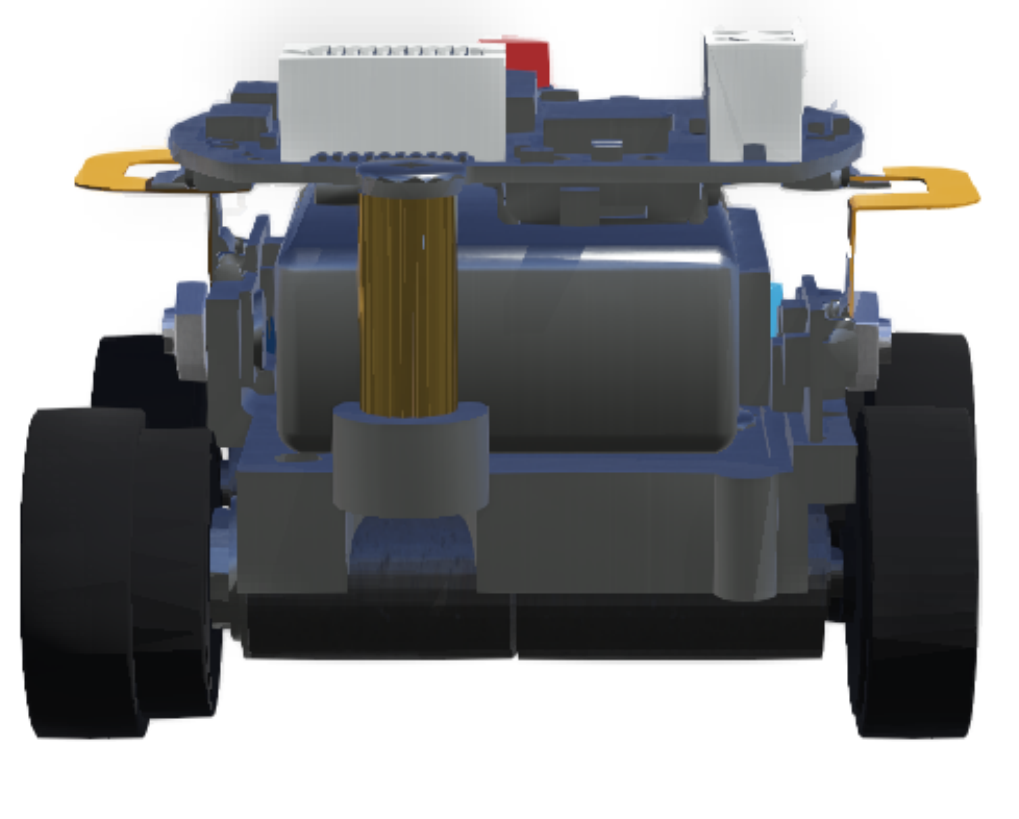}
        (a) Model, front
    \end{minipage}
    \begin{minipage}[t]{0.19\textwidth}
        \centering
        \includegraphics[width=0.9\textwidth]{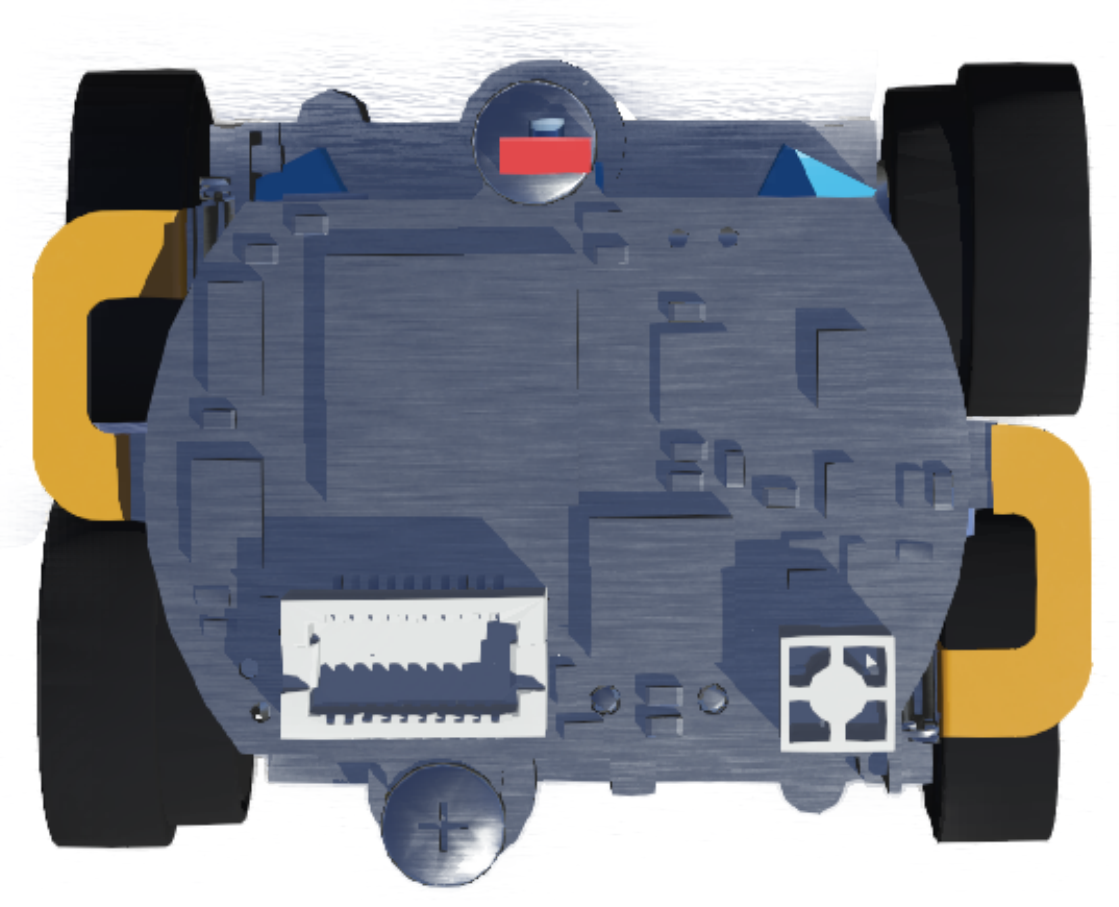}
        (b) Model, top
    \end{minipage}
    \begin{minipage}[t]{0.21\textwidth}
        \centering
        \includegraphics[width=\textwidth]{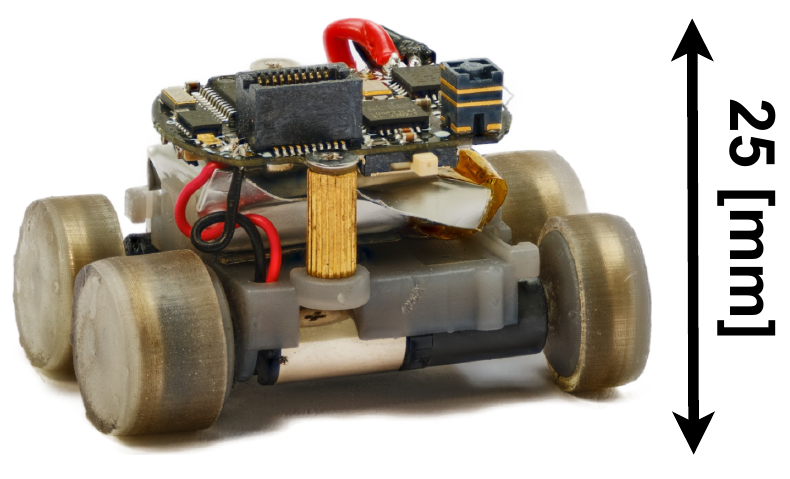}
        (c) Real, front
    \end{minipage}
    \begin{minipage}[t]{0.21\textwidth}
        \centering
        \includegraphics[width=\textwidth]{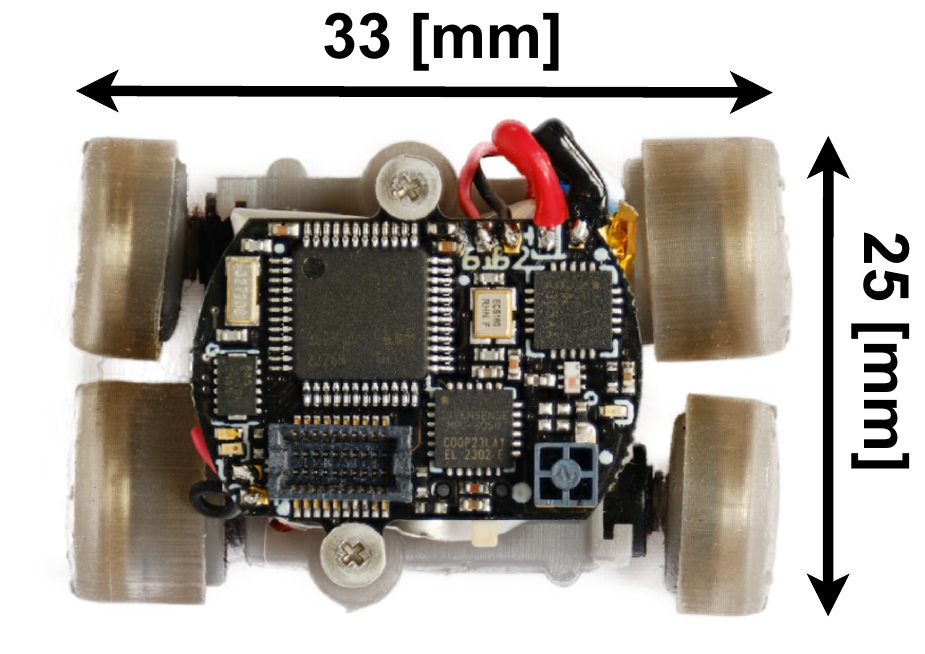}
        (d) Real, top
    \end{minipage}
\caption{We use a simpler robot model in simulation, with IR sensors directly simulated on the main PCB. Our simulated collision avoidance closely matches reality.}
\label{fig:robot_views}
\end{figure}

\section{Real Experimental Setup}Our experimental setup, shown in Figure \ref{fig:experimental_setup}, is built around (i) a tiled surface section of size $ 1\text{m}\times1 \text{m}$, and (ii) a swarm of $3 \text{cm}$-sized vibration-sensing wheeled robots that traverse and inspect the tiled surface section.
The surface section consists of 25 tiles, each of size $20 \text{cm}\times20 \text{cm}$, that are laid out in a square grid with five tiles on each side. There are two types of tiles on the surface, vibrating and non-vibrating tiles. The vibrating tiles are excited using two miniature vibration motors mounted on top of one another underneath the tile at its center (\footnotesize{ERM 3V Seeed Technology motors}). All tiles are secured to an aluminium frame using $2 \text{cm}\times1 \text{cm}$ pieces of magnetic tape around the corners. The frame consists of four strut profiles in the middle and four others along the edges of the arena. We use an overhead camera ({\footnotesize{Logitech BRIO 4k}}) and AruCo markers for visual tracking of the robots.

We use a revised and extended version of the Rovable robot originally presented in \cite{Dementyev2016Rovables:Wearables}. As shown in Figure \ref{fig:vibration_sensing_robot}, each robot measures $25\text{mm} \times 33\text{mm}\times 35\text{mm}$ and carries two customized Printed Circuit Boards (PCBs): (i) a main PCB hosting the micro-controller, an IMU, motor controllers, power circuitry and radio and (ii) an extension PCB hosting IR sensors for collision avoidance. The main PCB has essentially the same design as the one in \cite{Dementyev2016Rovables:Wearables}, and was only revised and remade for updated components. The extension PCB is new and hosts three small Time-of-Flight (ToF) IR sensor boards, each facing a direction of $0^{\text{o}}$, $25^{\text{o}}$, and $-25^{\text{o}}$ relative to the forward driving direction of the robot. Each sensor has a field of view of $27^{\text{o}}$ and a range of up to $1\text{m}$. A 3D printed shield is mounted around the extension PCB to enhance the visibility of the robot when perceived by the IR sensors of other robots. The robot has four magnetic wheels. Only two wheels, one on the front and one on the back, are driven by PWM operated motors. At 100\% PWM, the robot drives forward at around $5\text{cm}$ per second.

\section{Simulation and Optimization Framework}
\begin{figure}[t]
\centering 

    \begin{minipage}[t]{0.24\textwidth}
        \centering
        \includegraphics[width=\textwidth]{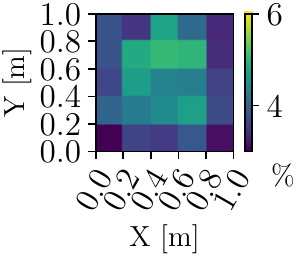}
        (a) Experiment
    \end{minipage}
    \begin{minipage}[t]{0.24\textwidth}
        \centering
        \includegraphics[width=\textwidth]{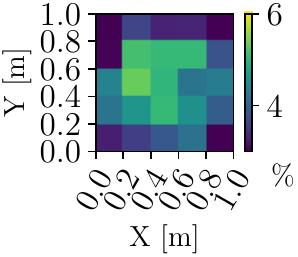} 
        (b) Simulation
    \end{minipage}
        \begin{minipage}[t]{0.24\textwidth}
        \centering
        \includegraphics[width=1\textwidth]{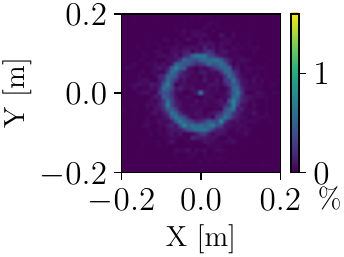}
        (c) Experiment
    \end{minipage}
    \begin{minipage}[t]{0.24\textwidth}
        \centering
        \includegraphics[width=\textwidth]{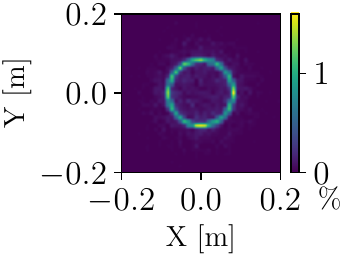}
        (d) Simulation
    \end{minipage}

    \caption{Distribution of samples across tiles in real (a) and simulated (b) setups. The displacement between consecutive samples in real (c) and simulated (d) setups.
    }
    \label{fig:sim_real correlation}
\end{figure}

Our simulation framework provides a virtual environment where we can study the operation of our robot swarm. 
Within Webots, we set up two main components: (i) a realistic model of our robot, and (ii) a tiled surface that the robots inspect, with a black and white projected floor pattern. We use the black and white tiles in simulation as a proxy for vibrating and non-vibrating tiles in our real experimental setup. In simulation, we assume noise-free binary sampling of the surface and zero loss on inter-robot communication. Figure \ref{fig:robot_views} shows the simulated and the real robots side by side. In simulation, the ToF sensor board is absent, but simulated IR sensors retain comparable range and positioning. We recreate mechanical differences that exist between real robots by adding randomized offsets to the simulated left and right motor speed commands:
\begin{subequations}
\begin{equation}
\label{eq:vel_motor_r}
    M_l^s \leftarrow M_l^s r_v (1- r_a)
\end{equation}
\begin{equation}
\label{eq:pso_pos} 
    M_r^s \leftarrow M_r^s r_v (1+ r_a)
\end{equation}
\end{subequations}
where $M_l^s$ and $M_r^s$ are the left and right motor speeds, and $r_v$ and $r_a$ are drawn from empirically chosen uniform distributions $U\sim(0.95,1.05)$ and $U\sim(-0.125,0.125)$, respectively.  
We calibrate our simulation empirically considering three characteristic features: (i) sample distribution over the experimental setup, (ii) time between consecutive samples, and (iii) distance between consecutive samples. To obtain data for our calibrations, we run experiments using five robots for 3 $\times$ 20 minutes with algorithm parameters $[\gamma,\gamma_0,\tau, \theta_c] = [5000,2000,1500,60]$ using the $u^-$ information sharing strategy (Section \ref{sec:algorithm}). The Pearson correlation coefficients for the obtained data (partly shown in Figure \ref{fig:sim_real correlation}) corresponding to the three features mentioned above are calculated as $0.990$, $0.984$, and $0.735$, respectively. These values confirm the similarity between simulated and real swarm behaviors, enabling optimizing real experiments using simulation.

Our optimization framework involves two components: (i) our calibrated simulation and (ii) a noise-resistant PSO method. Throughout the PSO iterations, every particle is evaluated multiple times on randomized floor patterns with the same fill-ratio. The velocity and position of particle $i$ are updated at iteration $k$ as:
\begin{subequations}
\begin{equation}
\label{eq:pso_vel}
     \mathbf{v}_{i}^{k+1} = \omega \cdot \mathbf{v}_{i}^k + \omega_p \cdot \mathbf{r_1}(\mathbf{p}_{b_{i}} - \mathbf{p}_{i}^k) + \omega_g \cdot \mathbf{r_2}(\mathbf{g}_{b_{}} - \mathbf{p}_{i}^k)
\end{equation} 
\begin{equation}
\label{eq:pso_pos}
    \mathbf{p}_{i}^{k+1} = \mathbf{p}_{i}^k + \mathbf{v}_{i}^{k+1}
\end{equation}
\end{subequations}
where $\mathbf{v}_{i}^k$ and $\mathbf{p}_{i}^k$ are the velocity and position vector of particle $i$ at iteration $k$. $\mathbf{p}_{b_{i}}$ and $\mathbf{g}_{b_{}}$ correspond to the position vector of the personal best and global best evaluations for particle $i$, respectively. We set the PSO weights for inertia, personal best, and global best as $[\omega \ \omega_p \ \omega_g ] = [0.75 \ 1.5 \ 1.5 ]$, balancing local and global exploration \cite{Poli2007ParticleOptimization,Gad2022ParticleReview,Innocente2010CoefficientsGuidelines}. The values $\mathbf{r_1}$ and $\mathbf{r_2}$ are drawn from  a uniform distribution $U\sim(0,1)$ each iteration.

\section{Experiments and Results}
We use a swarm of five robots, all employing $p_c=0.95$, and floor patterns with a fill ratio of $f=0.48$ to conduct simulation and real experiments. We evaluate on decision time and accuracy for the strategies $u^-$, $u^+$, and $u^s$. We consider decision time as the time the last robot that makes a final decision. We average the beliefs (Equation \ref{eq:belief}) of the robots at this decision time to calculate a corresponding decision accuracy. We do not incorporate any base-line method, as this was already shown in \cite{Ebert2020BayesSwarms}.

\subsection{Simulation Experiments}
\begin{table}[b]
\caption{The PSO optimization parameters and bounds. $P_0$ is the empirical best guess particle. $P^*$ is the resulting best particle with respect to our cost-function.}
\label{table:PSOparameters}
\centering
 \begin{tabular}{|l|l|l|l|l|l|l|l|}
\hline
\textbf{Parameter} &$\gamma_0$[ms]&$\gamma$[ms]&$\tau$[ms]&$\theta_c$[mm]&$\theta_o$ \\ \hline
$P_0$         &2000&5000&2000 &60&50 \\ \hline
$\text{min}_i$&2000&0&1000 &50&50\\ \hline
$\text{max}_i$&15000&15000&3000&100&200 \\ \hline
$P^*$&7565&15000&2025&50&85  \\ \hline

\end{tabular}
\end{table}
Five algorithmic parameters determine the sampling behavior of the swarm. These include mean ($\gamma_0$) and mean absolute deviation ($\gamma$) of the Cauchy distribution characterizing the robots' random walk, the sampling interval ($\tau$), the collision avoidance threshold ($\theta_c$), and the observations threshold ($\theta_o$). Table \ref{table:PSOparameters} lists the boundaries of our optimization search space. The lower bounds for $\tau$ and $\theta_c$ are set to allow smooth pause-sample-move and collision avoidance maneuvers. The bounds on $\gamma$ and $\gamma_0$ are set such that a robot is able to cross the arena in one random walk step. The bounds on the observation threshold $\theta_o$ are established empirically. The particles in the PSO swarm are initialized randomly within the bounded search space, with the exception of one particle $P_0$ set to an empirically chosen location.
Each particle is evaluated multiple times to mitigate randomness. For a particle $i$, we define the performance cost $\mathcal{C}_i$ as:
\begin{equation}
    \mathcal{C}_i = \mu \left( \begin{bmatrix}
        c_1&c_2& \hdots & c_{N_e}
    \end{bmatrix}^\top\right) + 1.1 \cdot \sigma \left( \begin{bmatrix}
        c_1&c_2& \hdots & c_{N_e}
    \end{bmatrix}^\top \right) 
    \label{eq:cost_function}
\end{equation}
where $N_e$ is the number of re-evaluations and $c_j$ is the outcome of the evaluation $j$:
\begin{subequations}
\begin{equation}
\label{eq:pso_pos}
    \epsilon_i(t,d_f) = \begin{cases}
        \epsilon_f \cdot t/T_{end} & d_f = d_f^* \\
        \epsilon_f \cdot \epsilon_d & d_f \neq d_f^* \\
        \epsilon_f & d_f =-1
    \end{cases}
\end{equation} 
\begin{equation}
\label{eq:pso_vel}
     c_j =\sum_{i=1}^{N_{r}} \epsilon_i
\end{equation}
\end{subequations}

\begin{figure}[t]
\centering
\begin{minipage}{0.5\textwidth}
    \centering
  \includegraphics[width=\linewidth]{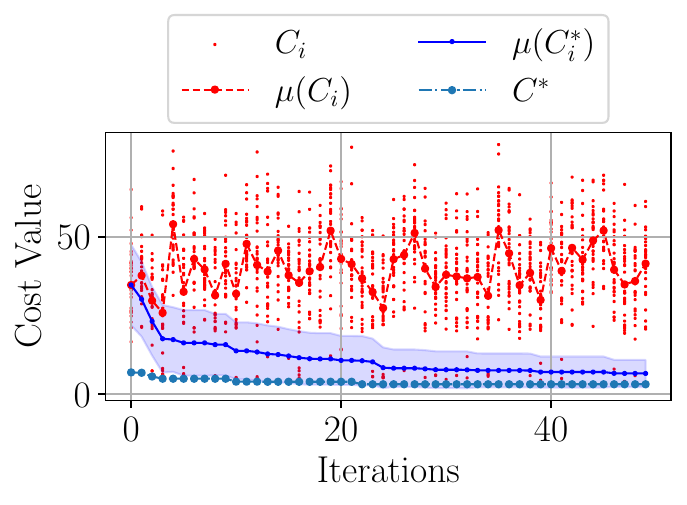}
  (a) Cost performance over iterations.
\end{minipage}%
\begin{minipage}{0.5\textwidth}
  \centering
  \includegraphics[width=\linewidth]{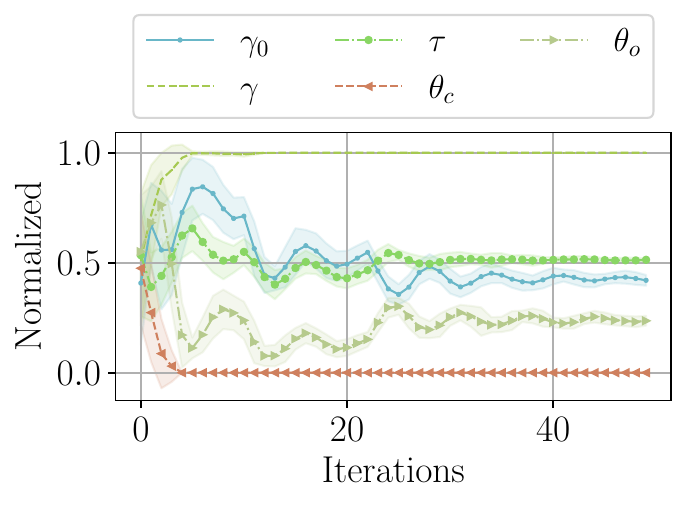}
  (b) Average values of particle dimensions.
\end{minipage}
\caption{
We use 30 particles, each re-evaluated 16 times, over 50 iterations using Equation \ref{eq:cost_function}. (a) Progression of cost performance of each particle ($C_i$), personal best cost performance of each particle ($C^*_i$), and the global best cost performance ($C^*$). (b) Progression of mean and standard deviation of parameters in Table \ref{table:PSOparameters}.}
\label{fig:PSO_results}
\end{figure}

where $N_r=5$ is the number of robots, $\epsilon_i$ is the performance cost of robot $i$ for which the robot's final decision $d_f$ made at time $t$ is compared with the correct decision $d_f^*$. A wrong decision is penalized by a factor of $\epsilon_d=5$. The value $\epsilon_f =1+|f-f^*|/\epsilon_t$ is calculated using the absolute difference between a robot's current estimate of the fill ratio $f=\alpha/(\alpha+\beta)$ and the correct fill ratio $f^*$, divided by a normalizing factor $\epsilon_t=0.04$ that corresponds to the contribution of one tile in the overall 25-tile setup. 

To find the optimal parameters for our inspection algorithm, we first consider running the algorithm with the $u^-$ information sharing strategy through our optimization framework. Our intuition is that an optimal parameter set for $u^-$ should allow the swarm to obtain a well-representative sample of the environment in a time-efficient manner, thus, the same parameters should also perform optimally for $u^+$ and $u^s$. Using this parameter set, we then run a systematic search to find an optimal value for the soft feedback parameter $\eta$ that characterizes the $u^s$ information sharing strategy.

We consider the $u^-$ strategy first. For the PSO optimizations, we use 30 particles, 50 iterations, and 16 re-evaluations. Each particle is evaluated for $T_{end}=1200s$ or until all robots in the swarm have reached a decision. The optimization results are shown in Figure \ref{fig:PSO_results}. It can be seen that the average personal best performance of the particles converges to the performance of the global best particle $P^*$, which is listed in Table \ref{table:PSOparameters}.

\begin{figure}[t]
\centering
\begin{minipage}{0.4\textwidth}
  \centering
  \includegraphics[width=\linewidth]{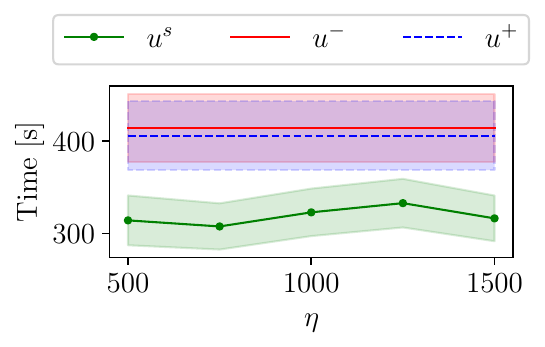}
  (a) Decision time for $\eta$ values.
\end{minipage}
\begin{minipage}{0.4\textwidth}
    \centering
  \includegraphics[width=\linewidth]{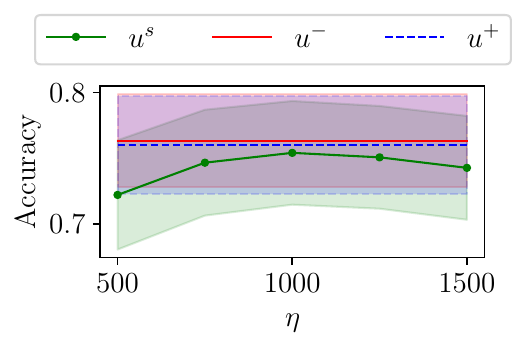}
 (b) Decision accuracy for $\eta$ values.
\end{minipage}
\begin{minipage}{0.4\textwidth}
    \centering
  \includegraphics[width=\linewidth]{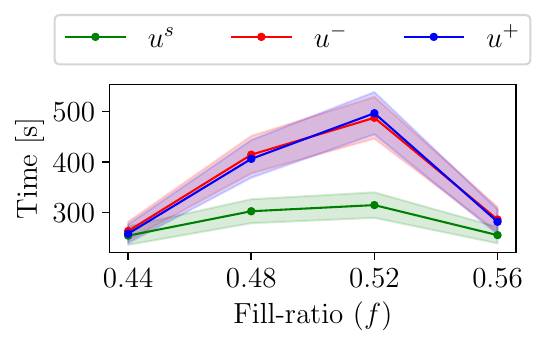}
  (c) Decision time for $f$ values.
\end{minipage}%
\begin{minipage}{0.4\textwidth}
  \centering
  \includegraphics[width=\linewidth]{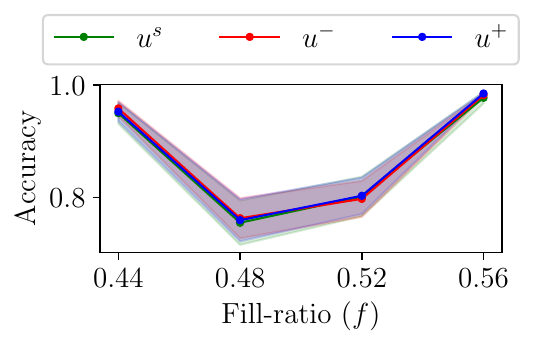}
  (d) Decision accuracy for $f$ values.
\end{minipage}
\caption{
Decision time (a) and decision accuracy values (b) of 100 randomized simulation experiments using systematic search for the soft-feedback parameter $\eta$ of $u^s$, compared against $u^-$ and $u^+$. Using $\eta=1000$ we run 100 randomized simulations for different $f\in [0.44,0.48,0.52,0.56]$ and compare with $u^-$ and $u^+$. The resulting decision times and accuracies are shown in (c) and (d), respectively.}
\label{fig:calibration_and_randomized_simulation}
\end{figure}

Using $P^*$, we then run a systematic search for the soft-feedback parameter $\eta$. We consider five candidate values for $\eta$ based on prior empirical tests and run 100 randomized simulations to evaluate the performance of $u^s$ against $u^-$ and $u^+$. Figure \ref{fig:calibration_and_randomized_simulation} illustrates the results. We see that $u^s$ consistently outperforms $u^-$ and $u^+$ in decision time. Regarding accuracy, $u^s$ closely approaches the performance of $u^+$ and $u^-$ at $\eta=1000$. Specifically, for $\eta=1000$, the $u^s$ achieves a $20.52\%$ reduction in mean decision time at a $0.78\%$ loss in accuracy, compared with $u^+$. When compared with $u^-$, $u^s$ achieves a reduction of $22.10\%$ in mean decision time at a $1.22\%$ loss in accuracy.

To assess the generalizability of our findings, we run 100 randomized simulations across fill-ratios of $f\in [0.44,0.48,0.52,0.56]$ to compare $u^-$, $u^+$ and $u^s$ (with $\eta=1000$) based on decision time and accuracy. Each simulation ends upon reaching $T_{end}$ or when all robots have reached a decision. For a fair comparison, we fix the random seeds used to generate floor patterns across the simulation instances. Figure \ref{fig:calibration_and_randomized_simulation}c shows that $u^s$ outperforms the other two strategies in decision time. Due to incorporating beliefs in messages, the swarm is compelled to make a decision rapidly, reducing mean and variation in decision times. This is particularly beneficial in harder environments where the fill ratio is close to $f=0.5$, facilitating reaching the credibility threshold $p_c$. 
\subsection{Real Experiments}
\begin{figure}[t]
\centering
\begin{minipage}[t]{0.24\textwidth}
    \centering
  \includegraphics[width=\linewidth]{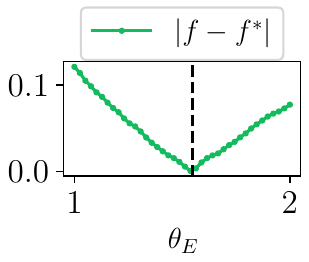}
  (a) $|f - f^*|$ vs $\theta_E$
\end{minipage}
\hfill
\begin{minipage}[t]{0.24\textwidth}
    \centering
  \includegraphics[width=\linewidth]{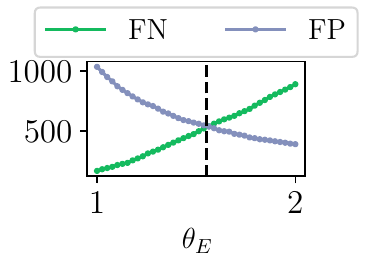}
 (b) FP $+$ FN vs $\theta_E$
\end{minipage}
\begin{minipage}[t]{0.24\textwidth}
    \centering
  \includegraphics[width=1\linewidth]{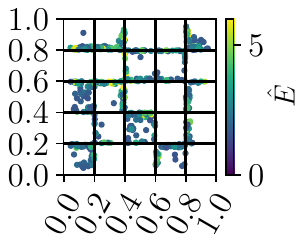}
 (c) FPs in arena
\end{minipage}
\begin{minipage}[t]{0.24\textwidth}
    \centering
    \includegraphics[width=1\linewidth]{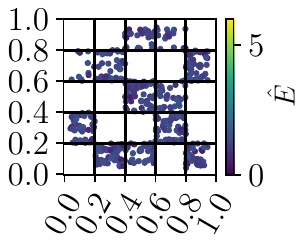}
 (d) FNs in arena
\end{minipage}
\caption{The threshold parameter $\theta_E$ determines binary observations of vibration data in real experiments (see Equation \ref{eq:observation}). (a) The fill ratio error $|f-f^*|$ for different values of $\theta_E$ (0.025 grid).  (b) False observations for different values of $\theta_E$ (c) Spatial distribution of false positive and (d) false negative observations.}
\label{fig:calibration_experimental_setup}
\end{figure}

We validate our simulation results by real experiments in 10 trials for $u^-,u^+$, and $u^s$.
We first tune the sample threshold $\theta_E$ using data from one hour of swarm operation employing $u^-$ and algorithm parameters $[\gamma,\gamma_0,\tau,\theta_c] = [5000,2000,1000,60]$, gathering a total of 6975 samples. Our evaluation criteria are the number of false observations and the fill-ratio error $|f - f^*|$. Figure \ref{fig:calibration_experimental_setup} shows that we obtain $|f - f^*| \approx0$ at $\theta_E=1.55$. Employing $\theta_E=1.55$ results in an equal amount of False Positives (FP) and False Negatives (FN), balancing the modeling error on the Beta distribution. Furthermore, we note that false observations appear mostly along edges of the tiles. This is expected as robots may sample close to the tile edges while they are in contact with two tiles.

\begin{figure}[t]
\centering
\begin{minipage}{0.4\textwidth}
    \centering
  \includegraphics[width=\linewidth]{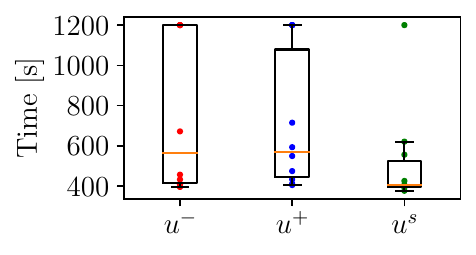}
  (a) Decision time for strategies.
\end{minipage}%
\begin{minipage}{0.42\textwidth}
  \centering
  \includegraphics[width=\linewidth]{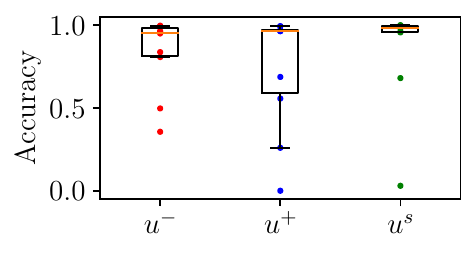}
  (b) Decision accuracy for strategies.
\end{minipage}
\caption{Decision time and accuracy in real experiments with a fill-ratio of $f=0.48$ for different information sharing strategies. If no decision is made within $t=T_{end}$, we refer to $T_{end}=1200${s} for decision time and the corresponding belief at the end of the simulation experiment for accuracy. We see that $u^s$ is faster and equally accurate as $u^-$ and $u^+$ in real experiments, confirming our findings in simulation experiments.}
\label{fig:experimental_trials_final_decisions}
\end{figure}

We conduct 10 experimental trials on our experimental setup with $f^*=0.48$ for assessing $u^-$, $u^+$ and $u^s$. 
The real experiments confirm our findings in simulation and reveal that the utilization of $u^s$ notably decreases the swarm's decision time. Employing $u^s$ compels the swarm to reach decisions at an accelerated rate compared to $u^+$ and $u^-$. The decision time and accuracy data from real experiments is shown in Figure \ref{fig:experimental_trials_final_decisions}. We can see that $u^s$ demonstrates inherently less variance in decision times. Moreover, we encounter fewer indecisive trial outcomes with $u^s$ compared to $u^+$ and $u^-$.

\section{Conclusion and Future Work}
In this work, we presented an experimental setup for studying a surface inspection task using a swarm of vibration sensing robots and explored the application of a Bayesian decision-making algorithm. We developed a simulation framework leveraging the physics based Webots robotic simulator and a PSO method to optimize the parameters shaping the robots' sampling performance. The resulting optimal parameter values were assessed for three information sharing strategies in randomized simulations across different environments based on the swarm's decision time and accuracy. We observed that our proposed soft feedback strategy yields a significant decrease in decision time without a major compromise in decision accuracy, compared to two previously studied strategies. Furthermore, hardware experimental trials validated our simulation findings. In real experiments, no drop in the decision accuracy was observed, demonstrating the adaptability and robustness of the decision-making processes to noise.
In our future work, we plan to increase the complexity of our experiments in several ways, considering (i) performing inspection of complex structures such as 3D surfaces or obstacle-dense environments, (ii) classifying time-varying fill-ratios on our experimental setup, and (iii) studying the effect of the swarm size on the inspection performance.

%
%
%
\bibliographystyle{splncs04}
\bibliography{references_ANTS}
%
%




%
\end{document}